\documentclass[letterpaper, 10 pt, conference]{ieeeconf}  

\usepackage[table]{xcolor}

\usepackage{multirow}
\usepackage{times}

\usepackage{hyperref}

\usepackage{color}
\usepackage{algorithm}
\usepackage{algpseudocode}
\usepackage{listings}
\usepackage{courier}
\usepackage{soul}
\usepackage{amsmath,amsfonts,amssymb}

\usepackage{mathtools}
\usepackage{subfigure}
\usepackage{wrapfig}
\usepackage{hyperref}
\usepackage{multirow}
\usepackage{graphicx}
\usepackage{graphicx}
\usepackage{soul}
\usepackage[flushleft]{threeparttable}
\usepackage{enumerate}
\usepackage{algorithmicx}
\usepackage{algpseudocode}
\usepackage{epstopdf}

\IEEEoverridecommandlockouts                              
\definecolor{red}{rgb}{1.00,0.00,0.00}
\definecolor{blue}{rgb}{0.00,0.00,1.00}
\definecolor{green}{rgb}{0.2,0.70,0.2}
\definecolor{yellow}{rgb}{0.5,0.5,0.0}
\definecolor{white}{rgb}{1,1,1}

\title{\LARGE \bf
Comparison Study of Well-Known Inverted Pendulum Models for Balance Recovery in Humanoid Robot
\author{Mohammadreza Kasaei, Nuno Lau and Artur Pereira
 \\IEETA / DETI University of Aveiro 3810-193 Aveiro, Portugal \\
	\{mohammadreza, nunolau, artur\}@ua.pt
}
}

\begin{document}
\maketitle
\thispagestyle{empty}
\pagestyle{empty}
\begin{abstract}
Bipedal robots are essentially unstable because of their complex kinematics as well as high dimensional state space dynamics, hence control and generation of stable walking is a complex subject and still one of the active topics in the robotic community. Nowadays, there are many humanoids performing stable walking, but fewer show effective push recovery under pushes.

In this paper, we firstly review more common used abstract dynamics models for a humanoid robot which are based on the inverted pendulum and show how these models can be used to provide walking for a humanoid robot and also how a hierarchical control structure could fade the complexities of a humanoid walking. Secondly, the reviewed models are compared together not only in an analytical manner but also by performing several numerical simulations in a push recovery scenario using \mbox{MATLAB}. These theoretical and simulation studies quantitatively compare these models regarding regaining balance. 
The results showed that the enhanced version of Linear Inverted Pendulum Plus Flywheel is the ablest dynamics model to regain the stability of the robot even in very challenging situations.

\end{abstract}
\vspace{-2mm}
\section{Introduction}
\label{sec:introduction}
Nowadays  number of researches in the field of humanoid robots is increasing and one of the common targets of these researches is realizing a humanoid robot which is able to operate in our dynamic daily-life environments with the same skill as humans. The application of this type of robot is not just doing our daily life tasks but also they can be used in several different applications such as rescue missions, helping incapable peoples, etc. Unlike wheeled robots, humanoid robot can adapt to our environments without facing limitations like gaps, uneven terrain and so on. It's just because of their similarity in kinematics as well as dynamics with a human. One of the essential requirements for using humanoid robots in such environments is the capability to perform tasks in a safe manner and the most important part of this requirement is stable locomotion. Generally, humanoid robots have more than 20 degrees of freedom~(DoF), therefore, they have complex dynamics as well as kinematic. In particular, they are unstable inherently, thus they need robust dynamics controllers to have mobility and robustness similar
to a human. During recent years, in order to develop a stable locomotion, several successful types of research have been introduced and can be generally divided into four categories: Central Pattern Generators~(CPG), passive dynamics control, heuristic-based methods and model-based methods ~\cite{picado2009automatic,7964063}. CPG methods are known as biologically inspired methods which try to design locomotion using generating some rhythmic patterns for each limb. Indeed, they are generally composed of several oscillators which are connected together in a specific arrangement. Passive dynamics methods describe the behaviors of robots by their passive dynamics and without using any sensors or control. These methods describe walking by considering the center of mass in pendulum falling until ground reaction forces redirect this motion into the next step cycle. Heuristic approaches (e.g., genetic algorithms, reinforcement learning, etc.) are generally based on learning methods. To have acceptable performance, these approaches require a lot of training samples. Thus, the learning phase in these approaches takes a considerable amount of time. These approaches are not commonly suitable to apply on a real robot due to the high potential of damaging the hardware during the learning phase~\cite{7964063}. All the above approaches are beyond the scope of this paper. The main focus of this paper is on model-based methods. 

In model-based approaches, a dynamics model of a robot is employed to generate reference trajectories of locomotion. In order to model the dynamics of robots, two different types of point of view exist. In the first point of view, the whole body dynamics (true model) of a robot is considered and in the second point of view, the overall dynamics of a robot is approximated by a simplified model. Although several significant achievements have been achieved based on both perspectives, a trade-off should be considered to select perspective. For instance, a true dynamics model can (not always) provide more accurate results but these methods are not only computationally expensive but also their results are totally platform-dependent. 
\begin{figure*}[t]
	\begin{centering}
		\begin{tabular}{c}
			\hspace{-0mm}
			\includegraphics[width=.95\textwidth, trim= 0cm 3cm 0cm 5.5cm,clip] {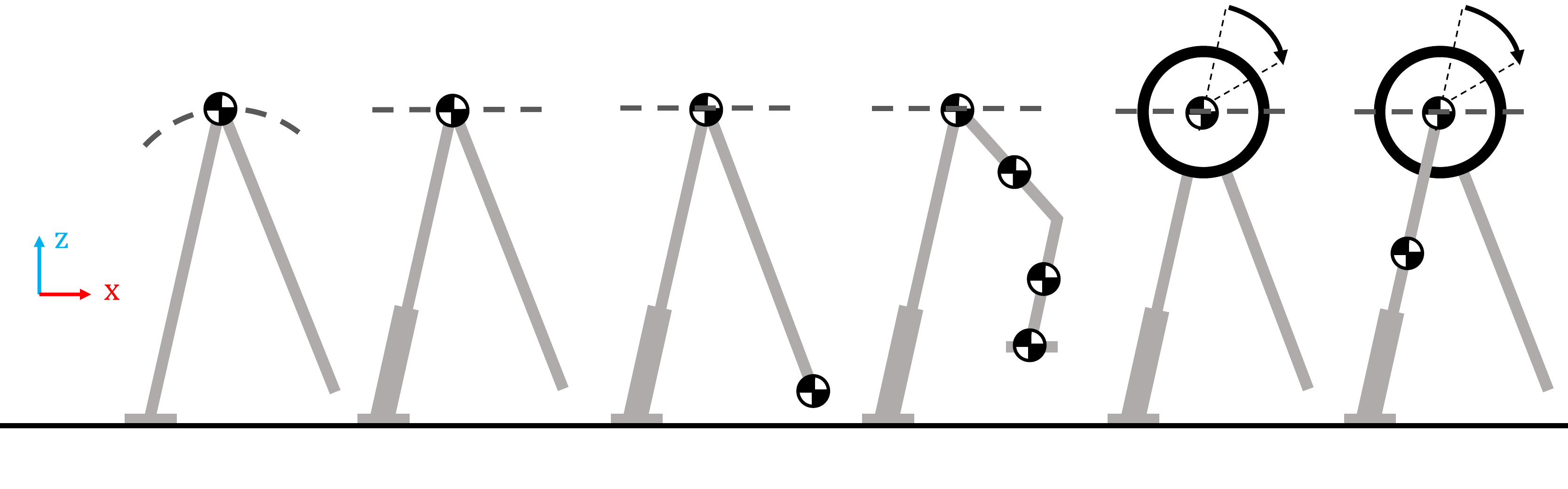}		
		\end{tabular}
	\end{centering}
	\caption{ Schematics of the simplified dynamics models presented in related work. In these schematics, gray links show the massless links which do not have any effects in the models. Black dashed lines indicate the trajectories of the COM. Name of the models from left to right are IP~\cite{hemami1977inverted}, LIPM~\cite{kajita1991study}, TMIPM~\cite{albert2003analytic}, MMIPM~\cite{albert2003analytic}, LIPPFM~\cite{pratt2006capture}, Enhance LIPPFM~\cite{kasaei2017reliable}.  }
	\label{fig:exp1}
\end{figure*}

The rest of this paper is focused on the second point of view to show how a simplified model provides insight into the fundamental principles of humanoid locomotion. Moreover, some well-known simplified dynamics models of humanoid robots are reviewed and compared together. 
This paper is structured as follows: Section~\ref{sec:related_works} gives an overview of related work. In the Section~\ref{sec:WalkingEngineStructure}, overall hierarchical architecture of a walking engine is presented, and each level of this structure are briefly explained. Moreover, formulation of the presented dynamics models are reviewed. Simulation results of the comparison are demonstrated in Section~\ref{sec:Simulation}. Finally, conclusions and future research are presented in Section~\ref{sec:CONCLUSION}.

\section {Related Work}
\label{sec:related_works}
The basic idea behind of using a simplified model instead of an exact one is organizing a complex system as a hierarchy. Generally, in hierarchical control approaches, a simplified model is used to determine the overall behaviors of the system in an abstract way and then by using a detailed
full-body inverse dynamics controller, these behaviors can be converted to individual actuator inputs \cite{faraji20163lp}. It's obvious that the performance of the system depends on the ratio of matching between the template and the exact model.

Several simplified models have been proposed and Inverted Pendulum~(IP)~\cite{hemami1977inverted} is one of those which is computationally efficient and straightforward to understand. This model describes
human dynamics in single support and provides a low-dimensional and physically-accurate model. In this model, the overall dynamics of a robot is approximated by a single mass which is connected to ground by a massless rod. 

Kajita and Tani~\cite{kajita1991study} restricted this model by defining a height constraint to the horizontal plane of the system. This simplification not only causes to reduce the computational cost but also provides an appropriate framework to control. Indeed, this constraint causes the dynamics of the system becomes completely first-order linear dynamic system. Later, Kajita et al.~\cite{kajita2001real} introduced the Three Dimensional Linear Inverted Pendulum Model~(3D-LIPM) and showed how this model can be used to generate walking in a 3D space. Afterward, In~\cite{kajita2003biped}, a preview control method based on Zero-moment point (ZMP) was designed to control the system.

Albert, et al.~\cite{albert2003analytic} proposed Two Masses Inverted Pendulum Model~(TMIPM) which is an extended version of LIPM that considers the mass of swing leg in order to increase the gait stability. In their method, trajectories of COM have been generated using a linear differential equation according to a predefined ZMP and swing leg trajectories. They extended their model by considering the dynamic
influence of the thigh, the shank and the foot of the swinging leg. Actually, the extended model is composed of four masses and it has been named Multiple Masses Inverted Pendulum Model~(MMIPM). Unlike LIPM, TMIPM and MMIPM do not have a direct solution because of dependency of motions of the masses to each other through the kinematic linkage, therefore, the authors proposed an iterative algorithm to define the trajectory of the torso. It should be noted that in their models, they considered the height of COM is a constant similar to LIPM.

Shimmyo, et al.~\cite{shimmyo2013biped} proposed another dynamics model which was composed of three masses which were located on the base link, the right leg, and the left leg. In order to use preview controller for generating the walking trajectories, they assumed two assumptions which were Constant Mass Distribution and Constant Mass Height. They showed the effectiveness of their method by experimental results. 

In all of the above models, the upper body is considered as a single mass, however the body of a humanoid robot has several DoF (i.e. waist, arms, and neck) and their motions can generate a momentum around the COM. If this effect is considered, the ground reaction force will not pass through the COM. As a consequence, if a proper method to manage these momentums is not considered, the robot could not keep its stability and may fall down~\cite{kasaei2017reliable}. To cope with this issue, some extensions to the LIPM have been proposed that considered the angular momentum around COM~\cite{pratt2006capture, komura2005feedback}. In~\cite{pratt2006capture}, the legs of the robot are considered to be massless and extensible. Besides, to model centroidal angular momentum about COM, a flywheel (also called a reaction wheel) is used instead of a point mass~(LIP Plus Flywheel Model or LIPPFM). According to this model, they proposed the capture point as well as capture region concepts which can be used to answer to this question: when and where to take a step while robot faces a massive magnitude push? Later, Stephens~\cite{stephens2007humanoid} used this model to determine decision surfaces that could describe when a particular recovery strategy (e.g., ankle, hip or step) should be used to regain balance.

In our previous work~\cite{kasaei2017reliable}, we proposed an enhanced version of LIPPFM and developed a reliable walking engine for biped robot based on this model. They released the height constraint
of the COM and showed how this enhancement allows a more human-like motion and more stable walking. Latter, In~\cite{kasaei2018optimal}, they extended their model by considering the mass of stance leg and showed the dynamics of the system could be represented using a first order differential equation by linearizing the model about the vertically upward
equilibrium. Besides, they showed how this model
could be used to plan and track the walking reference trajectories.

In the rest of this paper, a general hierarchical structure of biped walking will be presented and also we will explain how the presented models can be used to generate walking trajectories. Furthermore, the presented models will be compared together in a push recovery simulation scenario.

\section{Walking Engine}
\label{sec:WalkingEngineStructure}
\begin{figure}[!t]
	\label {StateMachine}
	\centering
	\includegraphics[scale=0.35, trim= 5cm 2.1cm 5cm 3.0cm,clip]{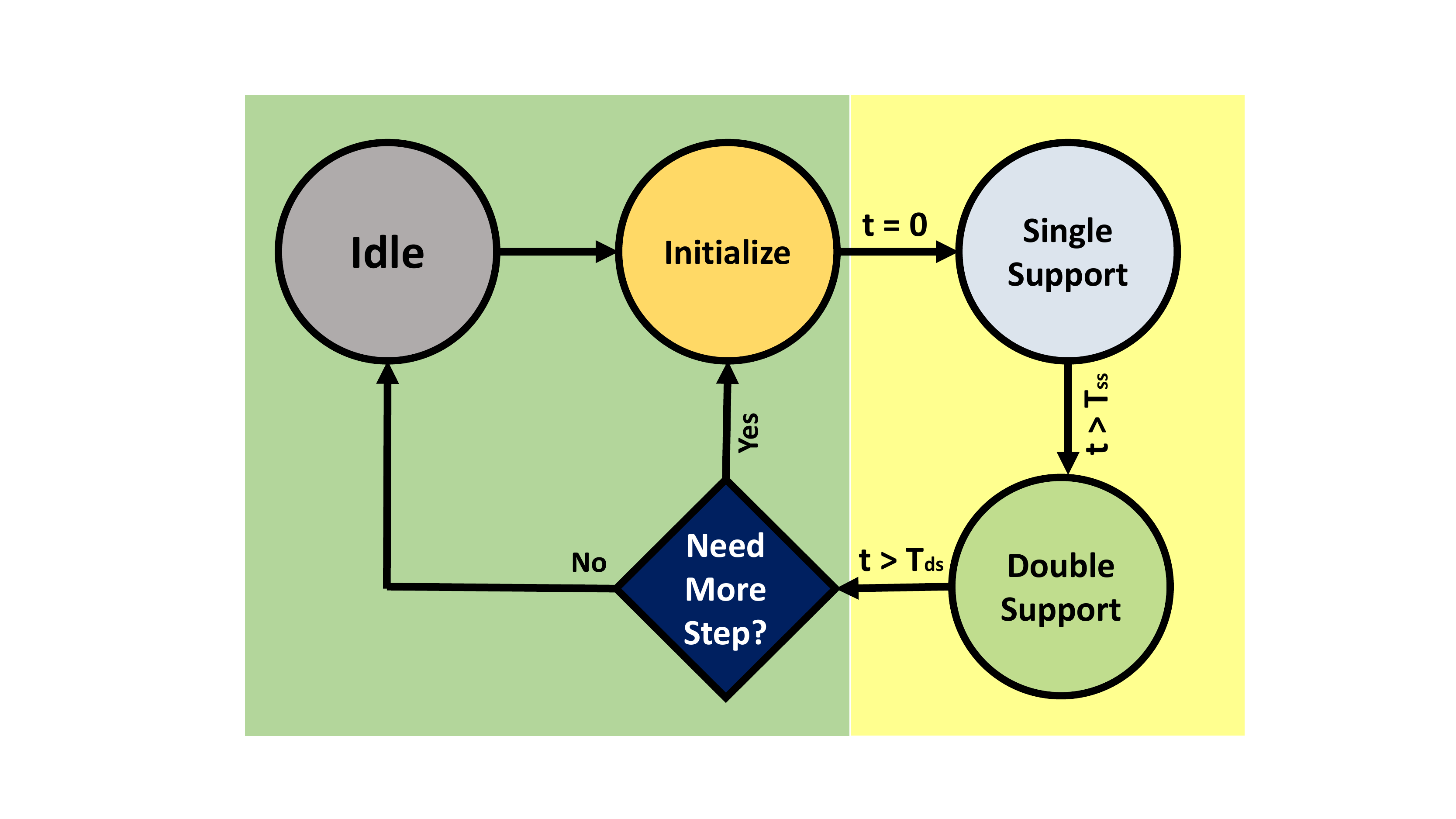}
	\caption{ Walking state machine with associated timer. Each state has a specific duration. }
	\label{fig:StateMachine}
\end{figure}

Walking is periodic locomotion which can be generated by repeating a series of steps and can be modeled using a state machine which is depicted in Fig.~\ref{fig:StateMachine}. As is shown in this figure, our walking engine is composed of four distinct states which are Idle, Initialize, Single Support and Double Support. In the Idle state, the robot is standing in place, and no walking trajectories are commanded. During Initializing state, the robot is going to be ready to start walking by moving its COM from between its stance feet to the first support foot. During Single Support as well as Double Support states walking trajectories has been generated and commanded. Moreover, a timer has been associated with this walking state machine to trigger a state transition. The timer increases $t$, and it will be reset once it reaches the duration of double support state. 

\begin{figure}[!t]
	\label {Hirachical}
	\centering
	\includegraphics[scale=0.42, trim= 8cm 1cm 4.5cm 1cm,clip]{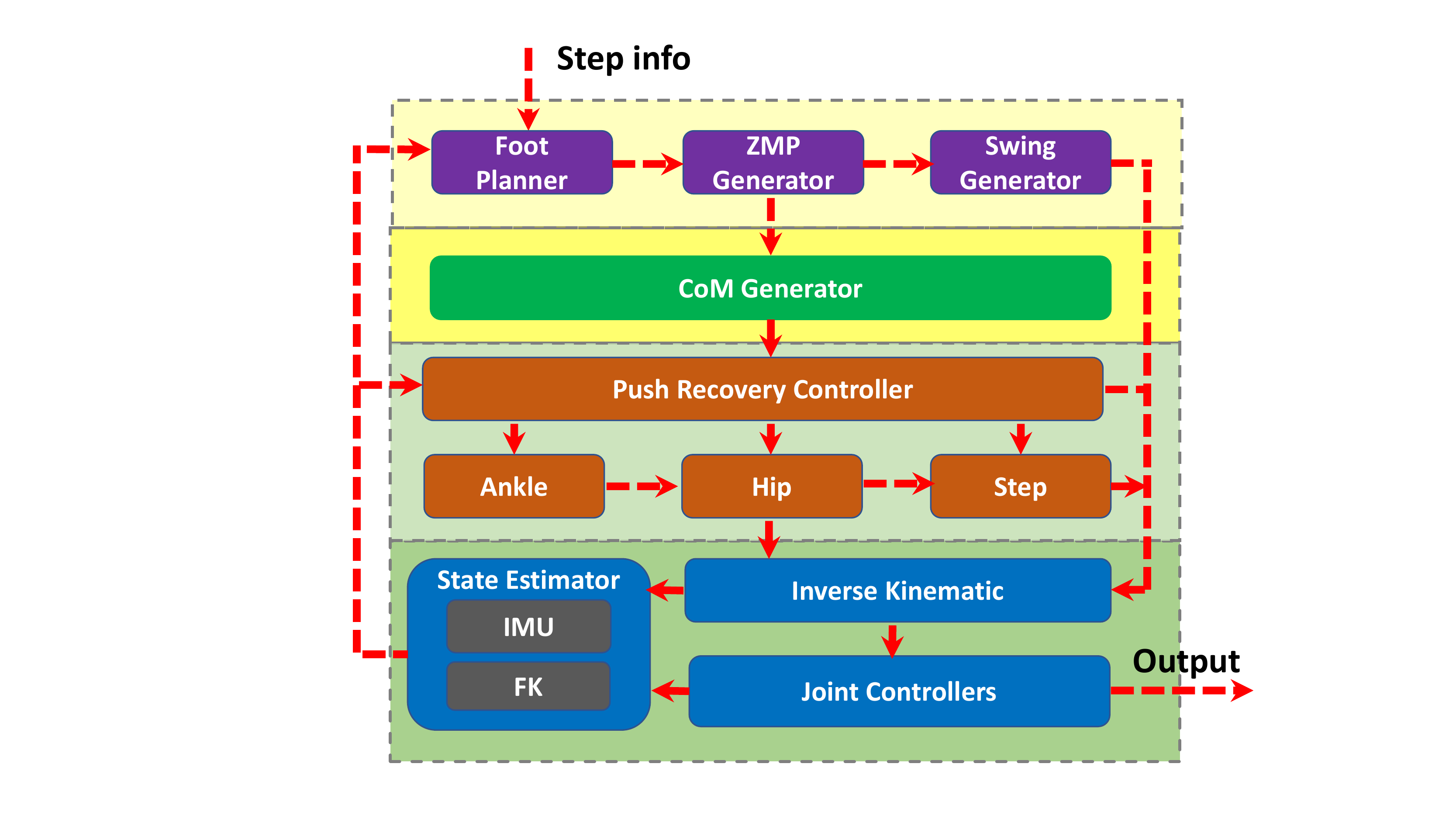}
	\caption{ Overall architecture of our hierarchical walking engine. It composed of four layers which is depicted by different colors.}
	\label{fig:Hirachical}
\end{figure}

In addition to this state machine, a hierarchical architecture is used to fade the complexities of the controller. The overall architecture of this controller is depicted in Fig.~\ref{fig:Hirachical}. As is shown in this figure, it is composed of four layers which will be described in the rest of this section.

\subsection{Foot Step Planner}
\label{Step_Planner}
This layer has three main tasks which are (i) generating a set of predefined foot positions (ii) generating the ZMP trajectories and (iii) generating the trajectories of the swing leg. All of these trajectories should be generated based on given step info and the predefined constraints (e.g., maximum step length, the minimum distance between feet, etc.). In our target, each step consists of two phases, single support, and double support and can be defined as follow:
\begin{equation}
\operatorname{Step} \equiv  \{L_{sx}, L_{sy} , T_{ss} , T_{ds}\}
\label{StepEq}
\end{equation}
\noindent
where $L_{sx}$, $L_{sy}$, $T_{ss}$, $T_{ds}$, represent step length, step width, single support duration and double support duration, respectively. These parameters should be selected based on the size of the robot, the capability of the robot and the tasks that the robot should perform. ZMP trajectories can be defined based on these parameters. The best intuitive choice for the ZMP trajectory during single support phase is the middle of the supporting foot, and it moves proportionally to the COM during double support phase. According to these assumptions, reference ZMP generator is formulated as follow:
\begin{equation}
r_{zmp}= 
\begin{cases}
\begin{cases}
f_{i,x} \\
f_{i,y} \qquad\qquad\qquad\qquad\qquad\qquad 0 \leq t < T_{ss} \\
\end{cases} \\
\begin{cases}
f_{i,x}+ \frac{L_{sx} \times (t-T_{ss})}{T_{ds}}   \\
f_{i,y}+\frac{L_{sy}\times (t-T_{ss})}{T_{ds}} \qquad\qquad\qquad T_{ss} \leq t < T_{ds} \\
\end{cases} 
\end{cases} ,
\label{eq:zmpEquation}
\end{equation}
\noindent
where $t$ represents the time which is reset at the end of each step ($t \geq T_{ss}+T_{ds}$), $T_{ss}$ , $T_{ds}$ are the duration of single and double support phases, respectively, $f_i = [f_{i,x} \quad f_{i,y}]$ is a set of predefined foot positions on a 2D surface ($i \in \mathbb{N}$). Thus, by determining these parameters and using Equation~\ref{eq:zmpEquation}, reference ZMP trajectories can be generated. 

After generating footsteps and ZMP trajectories, swing leg trajectories should be defined according to the dynamics model of the robot. In the case of considering mass less swing leg, these trajectories can be generated using arbitrary methods (e.g., polynomials, cubic spline, etc.). In other cases, these trajectories should be generated according to the dynamics model of the system.

\subsection{Gait Stability and COM Trajectories Generator}
Several criteria for analyzing the balance of a humanoid have been proposed and Zero-moment point~(ZMP) is one of the well-known approaches. Conceptually, ZMP is a point on the ground plane where the horizontal inertia and the gravity forces negate each other. Vukobratovic, et al.~\cite{vukobratovic1970stability} were the first ones that used ZMP as the main criterion to develop a stable walking for a humanoid robot.  In case of no external forces or torques the ZMP can be defined using the following equation:
\begin{equation}
p_x = \frac{\sum_{i=1}^{k} m_i x_i (\ddot{z}_i + g) - \sum_{i=1}^{k} m_i z_i \ddot{x}_i  }{\sum_{i=1}^{k} m_i (\ddot{z}_i + g)} \quad , 
\label{eq:zmp}
\end{equation}
\noindent
where $k$ represents the number of body parts which is considered in the dynamics model, $m_i$, $x_i$, $z_i$ represent the mass and positions of the $i_{th}$ body part. 

Generally, trajectories of COM are generated based on the dynamics model of the system and some predefined trajectories such as ZMP and also the swing leg in case of considering the mass of the swing leg. For some of the dynamics models presented in Section~\ref{sec:related_works}, an analytical solution exists to generate this trajectory (e.g., LIPM) and it can be counted as an important property of a dynamics model because it's not only straightforward but also computationally cheap. In other cases, where a direct solution is not feasible, these trajectories are generated based on some assumptions like~\cite{shimmyo2013biped} or it can be formulated as an optimization problem which is generally expensive regarding computation cost. In the rest of this subsection, COM trajectories generators of the  models presented in Section~\ref{sec:related_works} are briefly summarized.

\paragraph{LIPM}
 According to the Equation~\ref{eq:zmp} the dynamics model of LIPM is as follow:
\begin{equation}
\ddot{x} = \omega^2 ( x - p_x) \quad ,
\label{eq:lipm}
\end{equation}
\noindent
where $\omega = \sqrt{\frac{g}{z}}$ represents the natural frequency of the pendulum. This equation can be represented as a state space system:
\begin{equation}
 \begin{bmatrix} \dot{x} \\ \ddot{x} \end{bmatrix} = 
	\begin{bmatrix} 
		0 & 1   \\ 
	    \omega^2 & 0   
	\end{bmatrix}	 
	\begin{bmatrix} x \\ \dot{x} \end{bmatrix} +
	\begin{bmatrix} 0 \\ -\omega^2 \end{bmatrix} p_x \quad .
\label{eq:lipm_ss}
\end{equation}
\paragraph{TMIPM}
The dynamics of this model can be represented in state space form using the Equation~\ref{eq:zmp}:
\begin{equation}
\begin{aligned}
 \begin{bmatrix} \dot{x} \\ \ddot{x} \end{bmatrix} = &
\begin{bmatrix} 
0 & 1   \\ 
\omega^2 & 0   
\end{bmatrix}	 
\begin{bmatrix} x \\ \dot{x} \end{bmatrix} +
\begin{bmatrix} 0 & 0 \\ -\alpha & 1\end{bmatrix}\begin{bmatrix} p_x \\  \beta \end{bmatrix}
	\\&
\begin{cases}
\alpha = \frac{g}{z_c}+ \frac{m_c}{m_s\times z_c}(\ddot{z_s}+g)\\
\beta =  \frac{m_c}{m_s\times z_c} ( x_s\times(\ddot{z}_s + g) - \ddot{x}_s z_s)
\end{cases},
\end{aligned}
\label{eq:tmipm_ss}
\end{equation}
\noindent
where $x_s$, $z_s$ are the position of the swing leg in x and z-direction, $m_s$, $m_c$ represent the mass of swing leg and the remaining masses of the robot respectively. 

\paragraph{MMIPM}
The dynamics of the system is represented by the following differential equation:
\begin{equation}
 \ddot{x} = \omega^2 ( x - p_x)  + \underbrace{\sum_{i=1} \frac{m_i}{m_c \times z_c}\bigg((x_i-p_x)(g+\ddot{z}_i)-\ddot{x}_i z_i\bigg)}_{f(t)} ,
\label{eq:mmipm_ss}
\end{equation}
\noindent
where $z_c$ is the height of COM, $m_i$ represent the mass of $i_{th}$ part of the swing leg. As it was explained before, there is no direct solution for this model and, in such situations, the trajectories of COM should be generated using an iterative algorithm. Thus, for generating the COM trajectories, first, the system is assumed as a TMIPM and generate the trajectories of the COM using the Equation~\ref{eq:tmipm_ss} and predefined swing leg trajectories, then, based on a direct kinematic approach, the motions of $m_i$ are determined and then based on that motion $f(t)$ is calculated. This procedure executes until a $f(t)$ is found that satisfies the condition.

\paragraph{LIPPFM}
This model considers the momentum around the COM, and the equations of motion of this model can be represented using a first-order state space system as follow:
\begin{equation}
\begin{aligned}
\begin{bmatrix} \dot{x} \\ \ddot{x} \\ \dot{\theta} \\ \ddot{\theta} \end{bmatrix} = &
\begin{bmatrix} 
0 & 1 & 0 & 0  \\ 
\omega^2 & 0 & 0 & 0\\
0 & 0 & 1 & 0  \\ 
0 & 0 & 0 & 0  \\    
\end{bmatrix}	 
\begin{bmatrix} x \\ \dot{x} \\ \theta \\ \dot{\theta} \end{bmatrix} +
\begin{bmatrix} 0 & 0 \\ -\omega^2 & -(mL)^{-1} \\ 0 & 0 \\ 0 & I_w^{-1}\end{bmatrix}\begin{bmatrix} p_x \\ \tau_w \end{bmatrix},
\end{aligned}
\label{eq:LIPPFM}
\end{equation}
\noindent
where $m$ is the mass of flywheel, $L$, $I_w$ and $\tau_w$ represent the length of the pendulum, the rotational inertia of flywheel around flywheel center of mass and the flywheel torque respectively.
\paragraph{Enhanced LIPPFM}
In this model, the accuracy of the model has been improved by releasing the constraint on COM's height as well as considering the mass of pendulum in the model. The equations of motions can be represented in a state space form as follow:
\begin{equation}
\begin{aligned}
	\begin{bmatrix} \dot{\theta}_a\\ \ddot{\theta}_a \\ \ddot{\theta}_w \end{bmatrix}
	=&
	\begin{bmatrix} 
		0 & 1 & 0  \\ 
		\frac{\mu\times (g+\ddot{Z}_c)}{\gamma} & 0 & 0  \\
		\frac{-\mu\times (g+\ddot{Z}_c)}{\gamma} & 0 & 0  
	\end{bmatrix}	
	\begin{bmatrix} \theta_a\\ \dot{\theta_a} \\ \dot\theta_w \end{bmatrix}		
	+
	\begin{bmatrix} 
		0 & 0  \\
		\frac{1}{\gamma} & \frac{-1}{\gamma}  \\
		\frac{-1}{\gamma} & \frac{\gamma + I_w}{\gamma\times I_w}  
	\end{bmatrix}	
	\begin{bmatrix} 
		\tau_a  \\ \tau_w  
	\end{bmatrix} 
\\
&\begin{cases}
\gamma = M\times L^2 + I_p \\
\mu =  m\times l + M \times L
\end{cases},
\end{aligned}	
\label{eq:statespace_alpha}
\end{equation}
\noindent
where  $\theta = [\theta_a \quad \theta_w]^\top$ is a vector of pendulum and flywheel angles respecting to the vertical axis, $M$ and $m$ are the masses of flywheel and the pendulum, $L$ and $l$ are the lengths from the base of the pendulum to flywheel center of mass and to pendulum center of mass, respectively, $g$ describes the gravity acceleration, $\ddot{Z}_c$ represents the acceleration of COM in Z-direction, $I_p$ is rotational inertia of pendulum about the base of pendulum and $I_w$ represents rotational inertia of flywheel around  flywheel center of mass.
\begin{figure*}[!t]
	\begin{centering}
		\hspace{-4mm}
		\begin{tabular}	{c c}			
			\includegraphics[width=0.5\textwidth, trim= 1.9cm 0cm 2cm 0cm,clip]{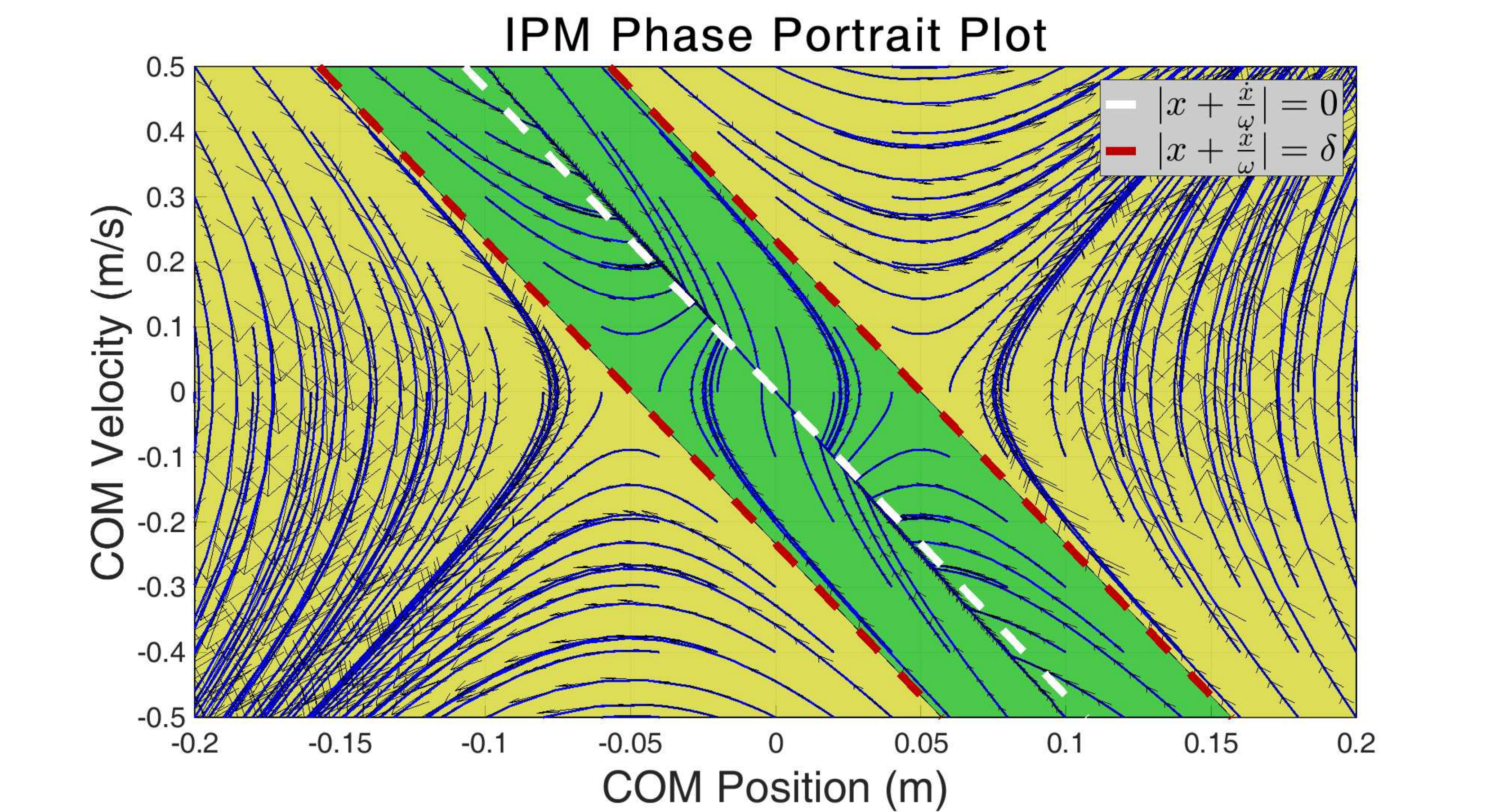} &
			\includegraphics[width=0.5\textwidth, trim= 1.9cm 0cm 2cm 0cm,clip]{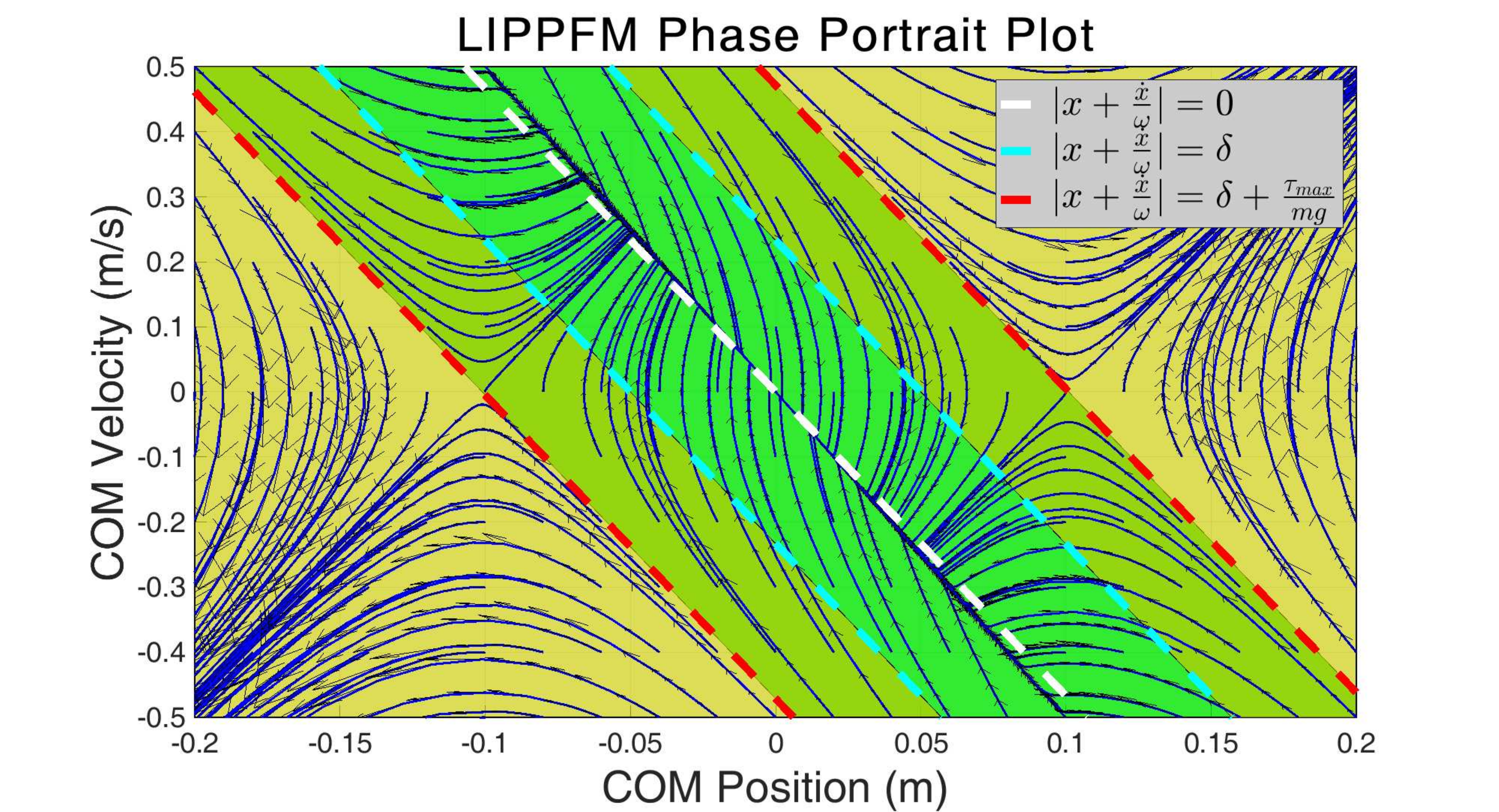} \\
			\includegraphics[width=0.5\textwidth, trim= 1.9cm 0cm 2cm 0cm,clip]{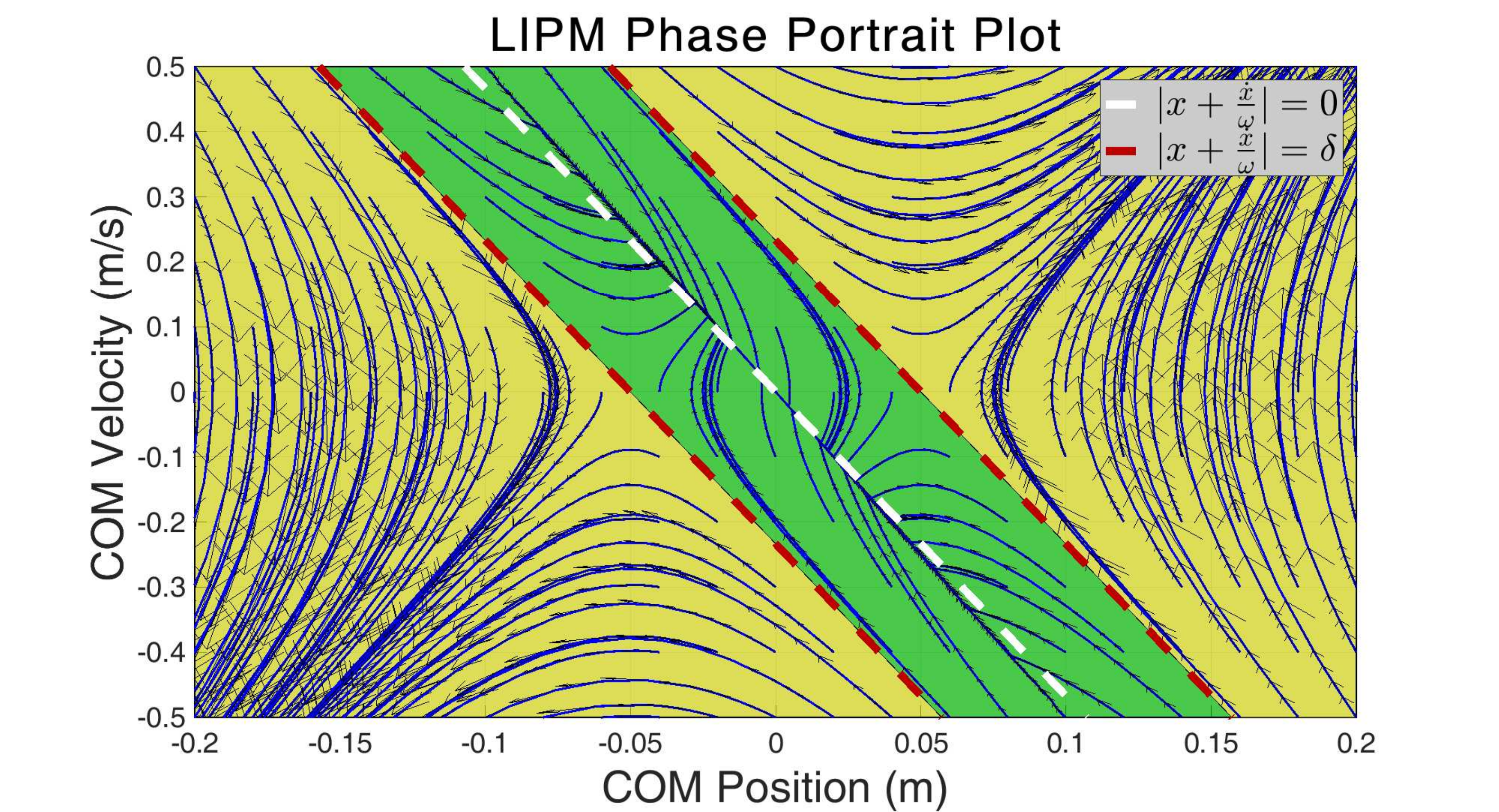} &
			\includegraphics[width=0.5\textwidth, trim= 1.9cm 0cm 2cm 0cm,clip]{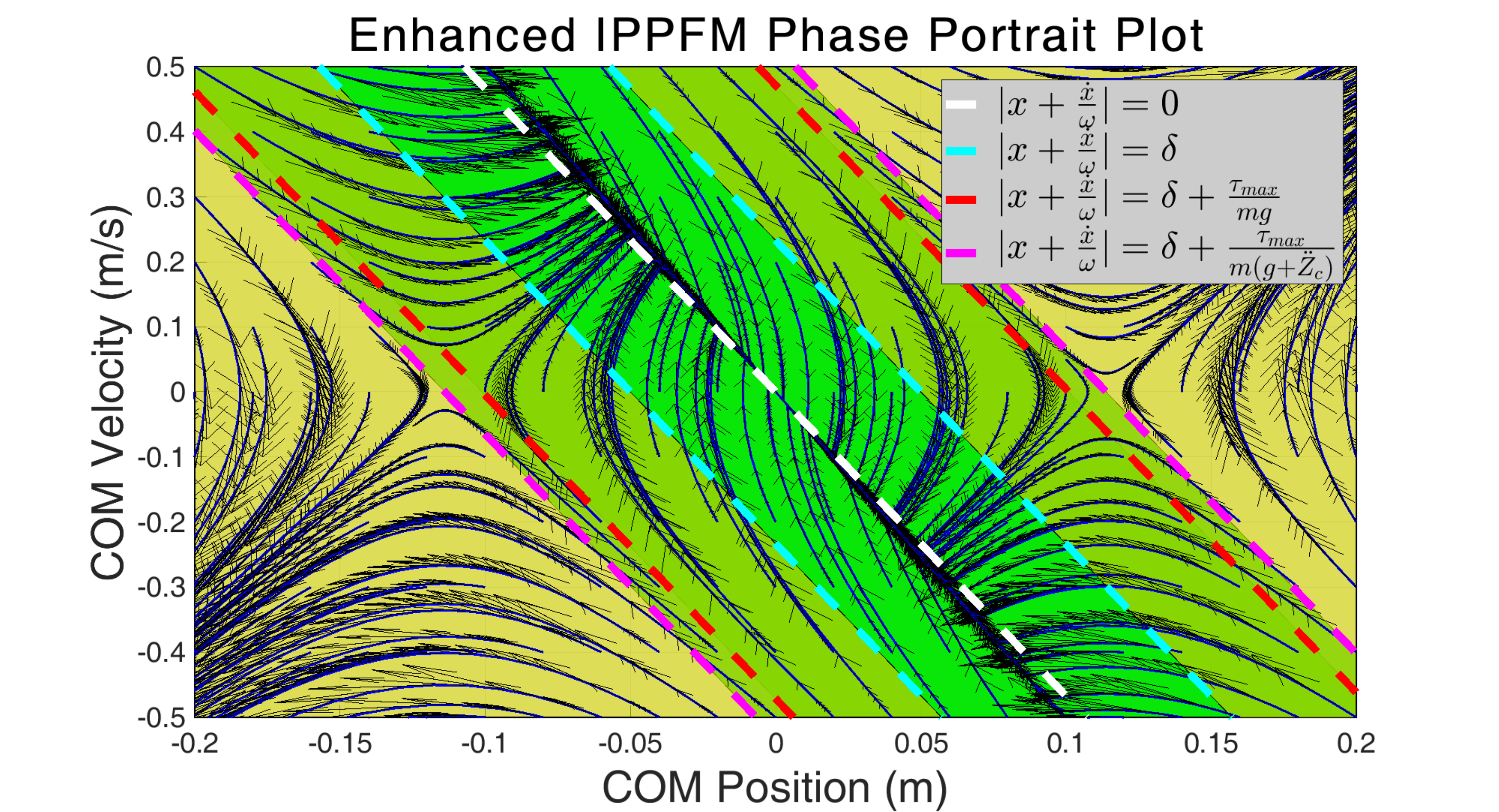} \\
			\includegraphics[width=0.5\textwidth, trim= 1.9cm 0cm 2cm 0cm,clip]{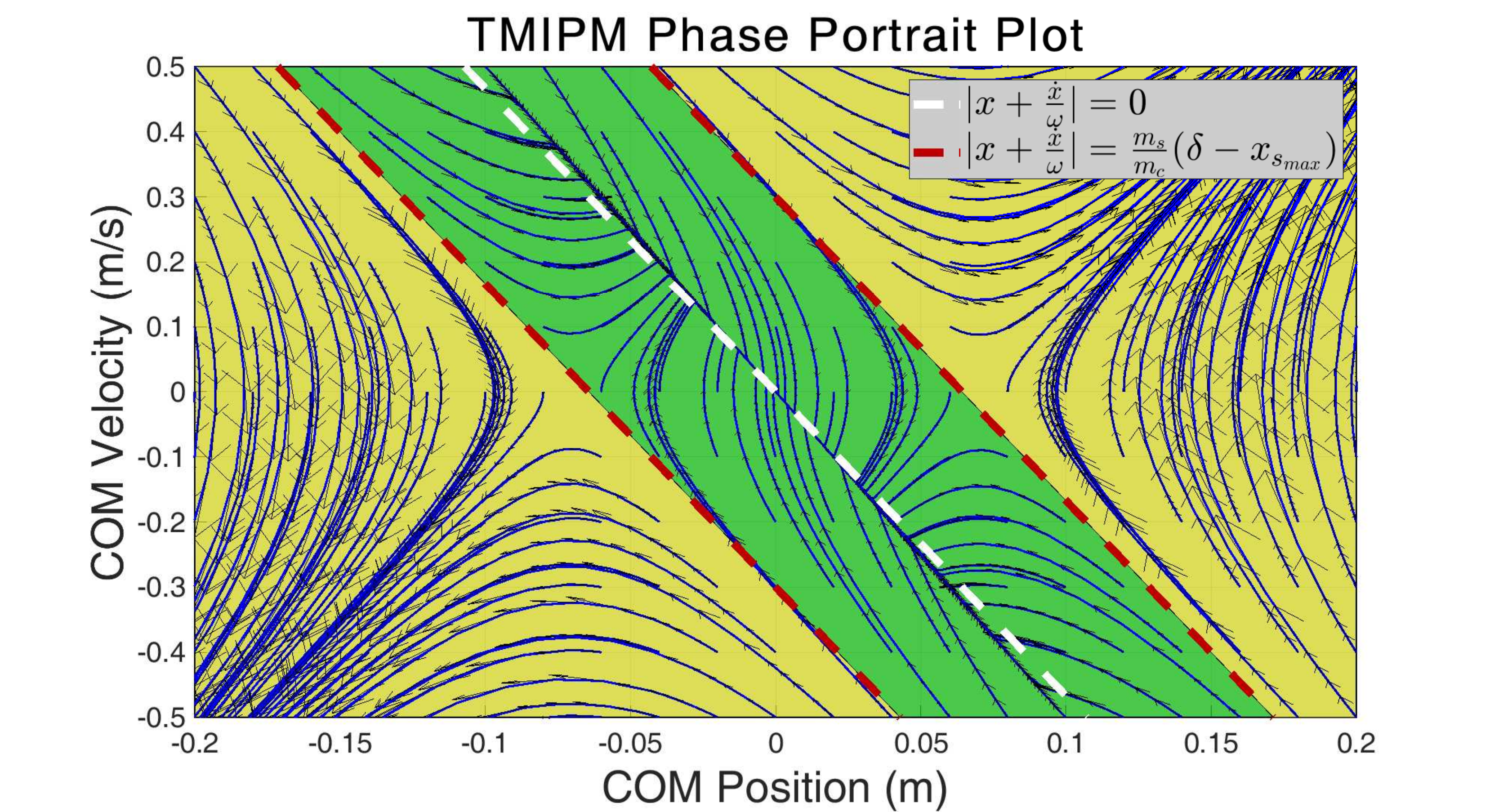} &
			\includegraphics[width=0.5\textwidth, trim= 1.9cm 0cm 2cm 0cm,clip]{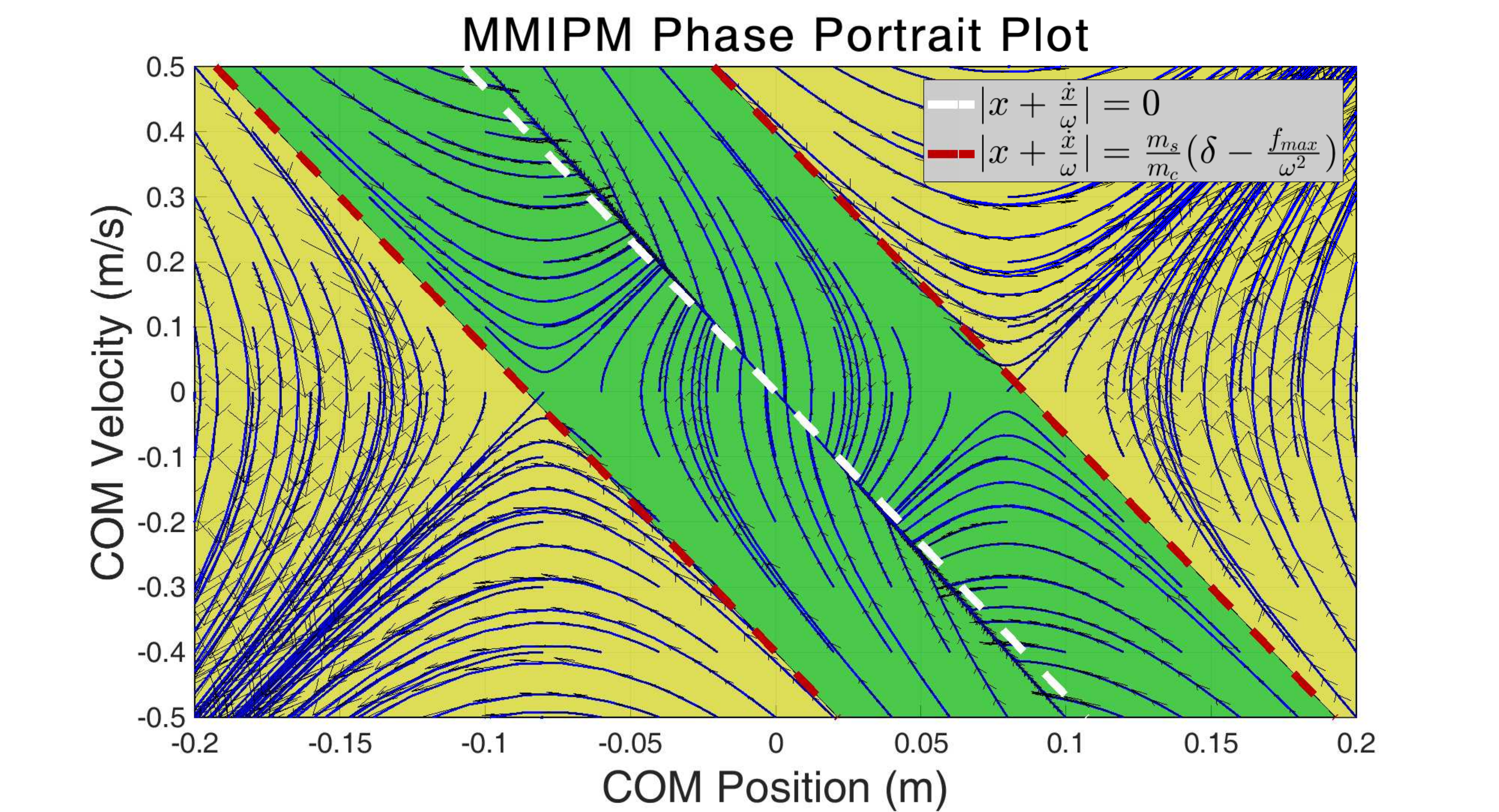} 
			
		\end{tabular}
	\end{centering}
	\caption{ Push recovery evaluation results. Yellow regions represent unstable regions where robot should take a step to regain its stability. Green (including light and dark) regions represent the stable regions which mean robot is able to regain its balance. Dash lines show the border of each recovery strategies.}
	\vspace{-5mm}
	\label{fig:push}
\end{figure*}
\subsection{Push Recovery Strategies}
A Feed-forward walking can be developed based on the trajectories of the ZMP and COM but this type of walking is not robust enough in facing with unexpected errors which can be raised from several sources, such as external disturbances, inaccurate dynamic model and etc. For instance, during walking on rough terrain environments, several forces will be applied to the robots. Hence, to keep the stability of the robot during walking, several criteria have been defined and the most important one is keeping the ZMP inside a polygon that is defined by the foot or feet touching the ground (support polygon). Human uses three distinctive actions including ankle, hip, and step recovery strategies to provide chances to regain balance. Ankle strategy tries to keep the stability of the robot by applying compensating torques at the ankle. Although this strategy improves the stability of the robot in some situations robot should use joints of the waist and the hips to prevent falling (hip strategy). In the case of significant disturbances, the stability of the robot can not regain using these strategies and robot should take a step. 

\subsection{Low Level Controller}
This level consists of three main modules including state estimator, inverse kinematics solvers, and joints position controller. The main task of this level is estimating position, velocity, and acceleration of the COM using IMU data that is mounted on torso or hip of the robot, and also a forward kinematic model of the robot which uses the values of the joint encoders.

\vspace{-1mm}
\section {Simulation Results and Comparison}
\label{sec:Simulation}
All of the presented models are able to generate walking for a humanoid robot in a general walking scenario but some of them are able to do it in a more stable manner. In order to compare the performance of these models, a push recovery simulation scenario has been defined. The goal of this scenario is examining the ability of models concerning regaining balance in different situations.
\begin{table}[h!]	
		\centering
	\caption{Parameters used in the simulations.}
\label{tb:parameters}
		\begin{tabular}{l|c|c|c|c} 
			\textbf{name} & \textbf{description} & \textbf{value}  & \textbf{min} & \textbf{max}\\
			\hline\hline
			$m_c$&mass ($kg$) & 7.00 &  &   \\
			$m_1$&mass of thigh($kg$) & 1.50 &   &   \\
			$m_2$&mass of shin ($kg$) & 1.50 &  &   \\
			$m_3$&mass of foot ($kg$) & 0.50 &  &   \\
			
			$Z_c$ & height of COM ($m$) & 0.45 & 0.40 & 0.50 \\
			$L_0$ & length of pendulum ($m$) & 0.50 & & \\			
			$L_1$ & length of thigh ($m$) & 0.28 & & \\
			$L_2$ & length of shin ($m$) & 0.28 & & \\
			$\delta$& length of foot ($m$) & 0.10 & & \\
			$\tau_w$& flywheel torque ($N/m$) & 0.00 & -5.00 & 5.00\\		
			$\ddot{Z}_c$& Acceleration in Z-direction ($m/s^2$) & 0.00 & -0.07 & 0.07\\		
			
		\end{tabular}
\end{table}
In these simulations, the robot is considered to be in single support phase and start from a specified initial condition $(x_0,\dot{x}_0)$ and robot should regain its stability without taking a step. According to the results of these simulations, we can find a specific answer for each model to this question: \textit{"when and which strategy(s) should be used to avoid falling?"}. Moreover, these numerical simulations allow the validation of the proposed formulations for each dynamics model. These simulations have been performed using MATLAB, and the most important parameters of the simulated robot as well as their ranges are shown in Table~\ref{tb:parameters}. For each dynamics model, a set of simulations have been run according to the set of initial parameters assumed for the simulated robot. The simulation results are depicted in the plots of Fig.~\ref{fig:push}. In these plots, each curve shows the result of a single simulation run. For each simulation, the simulated robot is started from single support with a specified initial condition and simulation. The initial condition is selected over the range of [-0.2 0.2] at interval $0.02 m$ for $x_0$ and [-0.5 0.5] at interval $0.1 m/s$  for $\dot{x}$ ( for each model 231 simulations were conducted ). Also, as it is shown in the plots, yellow regions show the unstable regions which mean robot could not keep its stability and green regions show stable regions which mean robot is able to regain its balance. 
\section {Conclusion}
\label{sec:CONCLUSION}
\begin{figure}[!t]
	\label {Summary}
	\centering
	\includegraphics[width = 0.9\linewidth, trim= 0cm 0cm 2cm 1cm,clip]{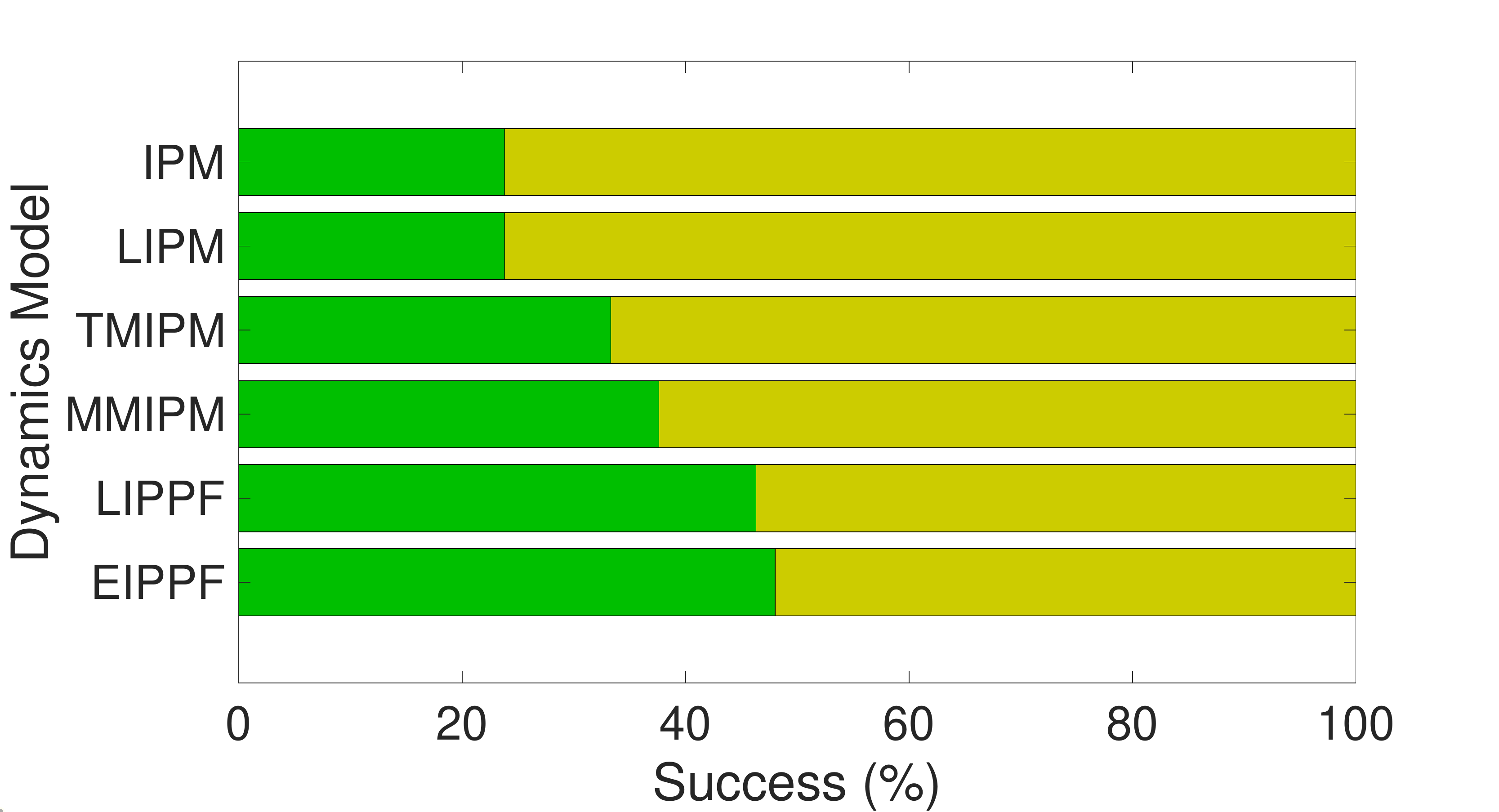}
	\vspace{-1mm}
	\caption{Summary of the simulation results.}
\vspace{-5mm}
	\label{fig:Summary}
\end{figure}

This paper presented a comparative study of some well-known dynamics model for balance recovery in a humanoid robot. This study was started by introducing some dynamics models and brief explaining how they generate the COM trajectories according to the input step parameters. Moreover, an overall architecture of a walking engine was presented to explain how the generated trajectories are used to produce walking. To validate the formulation and also compare them together, a set of simulations have been carried out using MATLAB. The results of the simulations are depicted in Fig.~\ref{fig:Summary}. As this figure shows, ELIPPFM is the ablest model to keep the stability of the robot even in very challenging conditions.
In future work, we would like to involve more complex dynamics models and also consider the stepping strategy in our comparisons. 

\section*{Acknowledgment}
This research is supported by Portuguese National Funds through Foundation for Science and Technology (FCT) through FCT scholarship SFRH/BD/118438/2016.

{
\bibliographystyle{IEEEtran}
\bibliography{Seminar}
}


\end{document}